\documentclass{isprs} 
\usepackage{subfigure}
\usepackage{caption}
\usepackage{setspace}
\usepackage{geometry} 
\usepackage{epstopdf}
\usepackage{multirow}
\usepackage[labelsep=period]{caption} 
\usepackage[british]{babel} 
\usepackage{url}

\begin{document}

\title{AN END-TO-END FRAMEWORK FOR LOW-RESOLUTION REMOTE SENSING SEMANTIC SEGMENTATION}

 \author{
  Matheus B. Pereira\textsuperscript{1}, Jefersson A. dos Santos\textsuperscript{1} 
  }

 \address{
     \textsuperscript{1} Dept. of Computer Science, Universidade Federal de Minas Gerais, Belo Horizonte, Brazil - \{matheuspereira, jefersson\}$@$dcc.ufmg.br
 }


\icwg{}   

\abstract{
High-resolution images for remote sensing applications are often not affordable or accessible, especially when in need of a wide temporal span of recordings. Given the easy access to low-resolution (LR) images from satellites, many remote sensing works rely on this type of data. The problem is that LR images are not appropriate for semantic segmentation, due to the need for high-quality data for accurate pixel prediction for this task. In this paper, we propose an end-to-end framework that unites a super-resolution and a semantic segmentation module in order to produce accurate thematic maps from LR inputs. It allows the semantic segmentation network to conduct the reconstruction process, modifying the input image with helpful textures. We evaluate the framework with three remote sensing datasets. The results show that the framework is capable of achieving a semantic segmentation performance close to native high-resolution data, while also surpassing the performance of a network trained with LR inputs.
}

\keywords{Super-resolution, semantic segmentation, remote sensing, end-to-end framework} 

\maketitle


\section{Introduction} \label{sec:introduction}

Remote sensing applications make use of satellite systems that sample many parts of the electromagnetic spectrum with dozens of spectral bands and with pixel sizes ranging from less than $1m$ to hundreds of meters \cite{schowengerdt2006}. Images with a high spatial resolution present more discernible textures and boundaries \cite{pouliot2018}, which helps many different computer vision tasks, such as semantic segmentation. Therefore, pixel resolution is an important factor that can highly impact how far a remote sensing application can perform as intended. Smaller resolutions make it difficult for pattern recognition algorithms to detect small objects and differentiate similar textures. High-resolution (HR) imagery is, then, desirable for most of the remote sensing applications.

The problem is that in practical scenarios, high-quality data is usually not accessible or affordable. HR satellite imagery is often expensive or not publicly available. They are also not present in a wide temporal range, which means that studies performed on older images usually can not rely on high-quality data. Drone imagery is difficult to obtain for large scale applications, such as the mapping of big areas. In urban environments, the use of drones is narrow due to restrictions imposed by law and the image acquisition requires more human intervention.

Due to the aforementioned problems, the use of LR satellite images is common in many remote sensing applications. LR satellites, such as Landsat 8, provide free (or cheap) world-wide data with a long history of acquisition. The main downside is the lack of spatial resolution that can compromise pattern recognition algorithms. Semantic segmentation is an example of a task that requires the most the input of HR images in order to perform accurately \cite{dai2016}.

Single-image super-resolution is a classic computer vision problem. The objective is to recover high-frequency details from a single low-quality input image. It is natural to expect that this problem could help to improve the results of different tasks when operating on LR data. A few works have studied this, especially for object detection \cite{haris2018task,ferdous2019,shermeyer2018}. For remote sensing semantic segmentation, to the best of our knowledge, only \cite{guo2019} and \cite{sibgrapi} evaluated the improvement with super-resolution. However, both of these works consider a case in which the semantic segmentation network is trained with HR data and tested on LR inputs. This case does not cover the situations in which we only have access to LR data for training. Furthermore, \cite{guo2019} and \cite{sibgrapi} use super-resolution as a pre-processing step for semantic segmentation, training the two tasks separately. In this work, we propose a framework that is trained in an end-to-end approach, which integrates a super-resolution and a semantic segmentation module. This allows the semantic segmentation to conduct the super-resolution training.

The main contribution of this paper is, therefore, the evaluation of a novel end-to-end framework that integrates super-resolution and semantic segmentation for the generation of high-quality thematic maps from LR inputs. We perform such evaluation on three distinct remote sensing datasets, comparing the results of the framework with a semantic segmentation network trained with native HR images and LR data.

The remainder of this paper is organized as follows. Section~\ref{sec:related} presents the works that studied the use of super-resolution for the improvement of different tasks. Section~\ref{sec:methodology} introduces our end-to-end framework and its technical aspects. Section~\ref{sec:experiments} details our experimental setup, while Section~\ref{sec:results} presents the results we obtained. Finally, Section~\ref{sec:conclusion} concludes the paper.

\section{Related Work} \label{sec:related}

Only a few works have studied the performance of semantic segmentation on LR data when applied with super-resolution. In \cite{dai2016}, the authors evaluated super-resolution for the improvement of edge detection, semantic segmentation, digit recognition, and scene recognition. Their overall conclusion was that super-resolution can improve the performance of those tasks for LR inputs. The main differences between their work and ours are: (i) they made only a superficial evaluation, employing methods that do not produce competitive results compared to the current state-of-the-art; (ii) their evaluation considers only everyday RGB images, and not remote sensing data; (iii) their evaluation is made by inputting super-resolved images into the semantic segmentation network trained with HR data, while our framework performs the whole training procedure with LR inputs, i.e. we do not have access to any HR input.

Haris \emph{et al.} proposed the Task-Driven Super-Resolution, which is the framework we base our end-to-end approach on. Their framework unifies super-resolution and object detection tasks in an end-to-end approach. They employed D-DBPN \cite{ddbpn} as their super-resolution module and SSD \cite{liu2016} as the object detection module. Their work includes the proposal of a loss function that unites the detection and reconstruction losses, which we will later adapt to the semantic segmentation task. Haris \emph{et al.} concluded that their framework with super-resolution can improve the performance of object detection compared to the use of LR inputs, and even when super-resolution is performed apart~\cite{haris2018task}. The main difference between our work to theirs is the fact that we evaluate the semantic segmentation problem, which requires more spatial information than object detection. Also, their framework was not tested with remote sensing data.

Some works employ super-resolution for object detection in low-resolution aerial imagery~\cite{ferdous2019,shermeyer2018}.
They employ SSD~\cite{liu2016} as the object detection method. 
Ferdous \emph{et al.} employed SRGAN \cite{ledig2017} for super-resolution and
Shermeyer \emph{et al.} chose VDSR \cite{kim2016vdsr}. They do not perform the training in an end-to-end manner. Their conclusion was also similar to the works previously cited. 

Two recent works have evaluated the use of super-resolution for the improvement of semantic segmentation on remote sensing imagery~\cite{sibgrapi,guo2019}. The first one selected D-DBPN \cite{ddbpn} as super-resolution network and Segnet \cite{segnet} as semantic segmentation network. 
The second approach uses ESPCN \cite{espcn} for super-resolution and U-Net \cite{unet} for semantic segmentation. While the first one manually degrades the training images, the method proposed by Guo \emph{et al.} inputs in the testing phase a panchromatic image from a different sensor of lower resolution.
Although similar to this paper, both of these works present two big differences. The first one is that they apply super-resolution as a pre-processing step for semantic segmentation, while our end-to-end framework trains both tasks at the same time. The second and most important difference is that they train the semantic segmentation network with HR images, while testing with LR ones. Despite being a valid situation in which we need to input an LR image into an already trained semantic segmentation network, this does not cover the cases in which we do not have access to HR data even for training. This is the case our paper approaches.

\section{Methodology} \label{sec:methodology}

\begin{figure*}[t]
\centering
\includegraphics[width=0.75\textwidth]{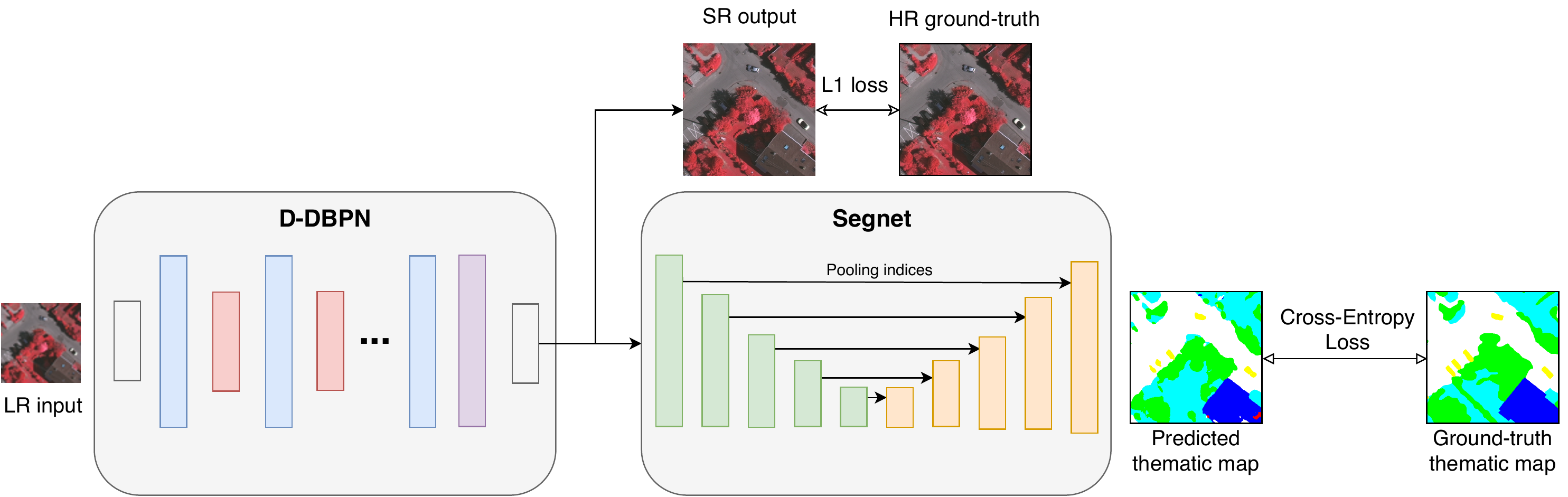}
\caption[]{Overview of the end-to-end framework, which trains the super-resolution and semantic segmentation networks at the same time. White blocks represent simple convolutional layers. Blue and red blocks represent, respectively, the up and down-projection units proposed in \cite{ddbpn}.The purple block is the concatenation of all the previous HR feature maps. Finally, the green and orange blocks represent, respectively, the encoder and decoder blocks proposed in \cite{segnet}.}
\label{fig:framework}
\end{figure*}

The proposed framework is mainly an adaptation of the task-driven approach introduced in \cite{haris2018task} for the semantic segmentation task. Our end-to-end framework can be divided into two modules: the super-resolution and the semantic segmentation. For the super-resolution part we employ the D-DBPN network, similarly to \cite{haris2018task} and \cite{sibgrapi}, while the semantic segmentation is performed with the use of Segnet \cite{segnet}. Figure~\ref{fig:framework} illustrates the framework, which performs the super-resolution before sending the result to the semantic segmentation network. First, the LR input image is sent to the framework, where it will first be processed by the super-resolution module. The result of this process will be a super-resolved image that will be used both to calculate the super-resolution loss and as input to the semantic segmentation module. After being processed by the Segnet, the final output of the framework will be an HR thematic map made from the LR input. This thematic map will also be used to calculate the semantic segmentation loss.

In the next subsections, we first present the super-resolution and semantic segmentation modules. Later, in Subsection~\ref{subsec:framework}, we discuss and give more details about the whole framework. We remark, though, that any other super-resolution or semantic segmentation deep-based network can be employed in the framework instead of the ones we selected.

\subsection{The Super-resolution Module} \label{subsec:sr}

D-DBPN, proposed in \cite{ddbpn}, was the chosen super-resolution network to be employed in the framework. This method gained attention for being the winner of the first track of the 2nd NTIRE challenge on single image super-resolution, which evaluated super-resolution methods using the bicubic downscaling with $8\times$ degradation factor. Being able to perform super-resolution with such a high degradation factor is the main reason why we chose to include this method in our framework.

The main characteristic of the D-DBPN network is the error feedback mechanism. D-DBPN sends the HR features back to the LR space using down-sampling blocks. This allows the network to guide the image reconstruction by calculating the projection error from many up and down-sampling blocks. The different ways of projecting back to another LR representation enriches the knowledge of the network, which learns various ways of up-sampling the features. There is a deep concatenation of HR features generated by all the previous up-sampling modules at the end of the network, while dense connections between the up and down-sampling blocks encourage feature reuse \cite{ddbpn}.

We use the same network configuration proposed in \cite{ddbpn}. We use $8 \times 8$ convolutional layers with four striding and two padding for $4\times$ super-resolution. For $8\times$ super-resolution, we use $12 \times 12$ convolutional layers with eight striding and two padding. Also following \cite{ddbpn} proposal, we set the number up and down-projection modules to 7.

For super-resolution, Peak signal-to-noise ratio (PSNR) serves as a metric to evaluate image restoration quality. We evaluate the PSNR over all the three channels of the inputs because two of the selected datasets are not RGB.  
The D-DBPN module is trained with the mean absolute error loss (L1).

\subsection{The Semantic Segmentation Module} \label{subsec:seg}

We employed Segnet~\cite{segnet} as the semantic segmentation module of the proposed end-to-end framework. This method has an encoder-decoder architecture and a pixelwise classification layer at the end. 
Each decoder layer has a corresponding encoder from which it
receives max-pooling indices to perform non-linear up-sampling of their input feature maps \cite{segnet}. This approach is less costly in terms of computational resources than using the full feature maps themselves, which is the main reason we selected this method. The final decoder output serves as the input of a softmax classifier to produce for each pixel the class probabilities.


The training of the Segnet is performed with the use of a pixelwise cross-entropy, which is also back-propagated to the super-resolution module. In order to evaluate the results of the semantic segmentation task, we selected the four metrics used in \cite{sibgrapi}: pixel accuracy, normalized accuracy, mean intersection over union ($IoU$) and Cohen's kappa coefficient ($Kappa$).


\subsection{The End-to-end Framework} \label{subsec:framework}

The end-to-end framework trains the super-resolution and the semantic segmentation network at the same time. This approach allows the semantic segmentation network to guide the super-resolution reconstruction in a way that is more beneficial for its own vision. When super-resolution is used as a separate pre-processing step, like in \cite{sibgrapi} and \cite{guo2019}, the super-resolution method does not take anything in consideration apart from the network's loss and the PSNR. Thus, we can say that the reconstruction is performed aiming to improve the image characteristics for a PSNR perspective only. The machine (semantic segmentation algorithm) vision, however, works differently from that. 
By allowing the semantic segmentation loss to be also used in the training procedure 
, we are letting it bias the reconstruction in a way that makes the image features more easily segmented.

The unified loss ($\xi$) of the framework is calculated as in Equation~\ref{eq:totalloss}, similarly to how \cite{haris2018task} applied it to the object detection task.

\begin{equation}
   \xi = \alpha L1(I_{HR}, SR(I_{LR})) + \beta Ce(y,Seg(SR(I_{LR}))),
    \label{eq:totalloss}
\end{equation}
where $L1(.)$ represents the super-resolution loss (mean absolute error) and $Ce(.)$ the cross-entropy loss for semantic segmentation. $I_{HR}$ and $I_{LR}$ represent, respectively, the HR ground truth image and the LR input. $y$ is the ground-truth thematic map. $SR(.)$ and $Seg(.)$ are, respectively, the result of the super-resolution and semantic segmentation modules. Finally, $\alpha$ and $\beta$ are pre-defined values that represent the balance between the super-resolution and semantic segmentation losses.

The definition of the $\alpha$ and $\beta$ values is the key to defining how biased the outputs will be for a PSNR or machine perception. With an $\alpha$ value higher than $\beta$, the network will prioritize the super-resolution reconstruction over the result of the semantic segmentation. However, by setting a $\beta$ value higher than $\alpha$, the framework will penalize more the semantic segmentation error and not care too much about how the image reconstruction is being performed. This means that the Segnet will be able to conduct the training of D-DBPN in a way it is better for the Segnet to see the relevant features. 

This framework is trained for $300$ epochs with inputs of size $480\times480$. The learning rate is initialized to $1e-5$ and is decayed by a factor of $10$ at half of the total epochs. The super-resolution module is optimized using Adam optimizer with $0.9$ momentum and $1e-4$ weight decay. The semantic segmentation module is also optimized with the Adam optimizer, but with $0.9$ momentum and $5e-4$ weight decay.

\section{Experimental Setup} \label{sec:experiments}
 
 In this section, we present the datasets and give details about the experiments.
 
\subsection{Datasets} \label{subsec:datasets}

In order to evaluate our framework, we selected the three remote sensing datasets used in \cite{sibgrapi}:

\begin{enumerate}
    \item Brazilian Coffee Scenes Dataset (\textbf{Coffee Dataset}) \cite{coffee}: This is a binary dataset (coffee and non-coffee classes). It contains images from three Brazilian cities:  Monte Santo, Guaran\'esia and Guaxup\'e (from the state of Minas Gerais). The images present green, red, and near-infrared bands. The dataset contains high intraclass variance caused by different crop management techniques and spectral distortions. Data from this dataset requires a lot of texture information in order to be accurately segmented.
    
    \item \textbf{Vaihingen Dataset}: Provided by the International Society for Photogrammetry and Remote Sensing (ISPRS) Commission for the 2D Semantic Labeling Contest, this dataset presents near-infrared, red and green bands. It contains six classes: impervious surfaces, building, low vegetation, tree, car, and clutter/background. This dataset contains small objects, such as cars, that will serve well to evaluate the framework's robustness.
    
    \item \textbf{Thetford Dataset}: The 2014 IEEE GRSS Data Fusion Contest dataset, which contains RGB sub-images from an urban area near Thetford Mines, in Quebec
    (Canada), and seven thematic labels: trees, vegetation, road, bare soil, red roof, gray roof, and concrete roof.
\end{enumerate}

We follow the same training protocol applied in \cite{sibgrapi}. For the coffee dataset, we train the framework on the images from the cities of Guaxup\'e and Montesanto, while testing on the images of Guaran\'esia. As for the Vaihingen dataset, we trained and tested our framework using only the publicly available images that contained the labeled ground-truth for semantic segmentation (16 images total).
As in \cite{nogueira2019} and \cite{sibgrapi}: areas $11$, $15$, $28$, $30$ and $34$ are used for testing, while the rest is used for training. We also exclude from the results the clutter/background class, like in \cite{nogueira2019} and \cite{sibgrapi}. For the Thetford dataset, we follow the proposal of the contest in terms of training and testing division. We exclude the bare soil class, as it is only present in the training part of the dataset.

\subsection{Experiment Details} \label{subsec:implementation}

We applied the same experimental protocol for each one of the datasets, following the approach proposed in \cite{sibgrapi}. Thus, we divide the HR images in crops of size $480 \times 480$. From the HR crops, we create the LR inputs with the use of bicubic interpolation (with $4\times$ and $8\times$ degradation). We initialize the weights of D-DBPN with the pre-trained model provided by \cite{ddbpn} in their official Github repository. Segnet weights are initialized with the VGG16 trained parameters for image classification.

We experimented different losses configurations for the end-to-end framework by changing the $\alpha$ and $\beta$ values of Equation~\ref{eq:totalloss}. As our objective is to improve semantic segmentation results, we aim for higher $\beta$ values. We tested the framework on the Vaihingen dataset with $\alpha$ values from the set $\{0.001, 0.01,$ $0.1, 1\}$, and $\beta$ from the set $\{1, 10, 100, 1000, 10000, 100000\}$. We observed that lower $\beta$ values achieved results that were similar to training super-resolution and semantic segmentation separately. 
We also observed that the higher the $\beta$ values, the higher was the number of artifacts created in the reconstructed image. Under these circumstances, the best results were achieved with the $0.1$/$1000$ configuration for $\alpha$ and $\beta$, respectively. Thus, the results reported next for the end-to-end framework are all using this same configuration.

The experiments were conducted in Python 3.6, with Pytorch 1.2 and a GTX Titan X 12GB. Due to the size of the framework, we can only execute it in one GPU with batch size $1$, thus the results reported in the next section all follow this configuration. The code for the proposed framework is available at~\url{https://github.com/matheusbarrosp/sr-semseg-end2end}.

\section{Results and Discussion} \label{sec:results}

The experiments were conducted in order to answer the following research questions:
(1) How robust is the end-to-end framework for different levels of degradation for remote sensing semantic segmentation and how does it compare to native HR data?
(2) How does the end-to-end framework compare to a semantic segmentation network trained with only LR images?
(3) Is the framework able to improve the segmentation accuracy of small objects?

\subsection{Robustness to different degradation levels}

In this subsection, we present the results that allow the evaluation of the robustness of the end-to-end framework with $4\times$ and $8\times$ degradation factor compared to the native HR data ($1\times$). 

\begin{table}[h!]
 \renewcommand{\arraystretch}{1.3}
 \centering
 \resizebox{\columnwidth}{!}{
 \begin{tabular}{llllll}
 \hline
 \textbf{Dataset}                    & \textbf{Deg.}           & \textbf{Acc}    & \textbf{Norm. acc} & \textbf{IoU}  & \textbf{Kappa}  \\
 \hline
 \multirow{3}{*}{Coffee}    & $8\times$  & 0.8003 & 0.7784 & 0.6477  & 0.5653 \\
                           & $4\times$  & 0.8205 & 0.8093 &  0.6807 & 0.6162 \\
                           & $1\times$ & 0.8330 & 0.8168 &  0.6972 & 0.6390 \\
 \hline
 \multirow{3}{*}{Vaihingen} & $8\times$  & 0.8288 & 0.6625 & 0.5654 & 0.7730 \\
                           & $4\times$  & 0.8293 & 0.6631 & 0.5697 & 0.7738 \\
                           & $1\times$  & 0.8479 & 0.6833   & 0.5909 & 0.7984 \\
 \hline
 \multirow{3}{*}{Thetford}  & $8\times$  & 0.8605 & 0.8564 & 0.6986 & 0.7881 \\
                           & $4\times$  & 0.8733 & 0.8417 & 0.7117 & 0.7997 \\
                           & $1\times$ & 0.8452 & 0.8184   & 0.6463 & 0.7636 \\
 \hline
 \end{tabular}
 }
 \caption{Semantic segmentation performance of the end-to-end framework for different degradation factors}
 \label{tb:semantic_segmentation2}
 \end{table}

Table~\ref{tb:semantic_segmentation2} shows the semantic segmentation results of the end-to-end framework. It is possible to see that the results for $8\times$ degradation factor while using the end-to-end approach are not far from $4\times$ degradation. 
In the Thetford dataset, the normalized accuracy was even higher when inputting $8\times$ degraded images compared to $4\times$, while the remaining metrics are also close. This indicates that the end-to-end framework is capable of dealing with higher degradation factors without losing too much accuracy. This is due to the fact that this framework can change the reconstructed image with information that is more easily discernible for the semantic segmentation task. When relying only on the super-resolution loss -- as it was done in \cite{sibgrapi,guo2019}, there is no assurance that similar textures (such as trees and vegetation, building and impervious surface) will be reconstructed in a way that highlights the differences among them. By letting the semantic segmentation task guide the super-resolution, we are allowing this highlighting to occur automatically. 

We can also see that even the difference to native HR data ($1\times$ degradation) is small in the proposed end-to-end framework. For the Coffee dataset, for example, the normalized accuracy was reduced from $0.81$ (for native HR data) to $0.77$ with $8\times$ degradation. This is only a $4\%$ difference from the original HR data to the restored one with the highest degradation factor.
The most interesting and noticeable change, though, is in regard to the Thetford dataset. The end-to-end framework actually managed to achieve better results with LR images than the Segnet trained with HR data. One of the reasons that made this happen is the low amount of training data for this dataset. The lack of training images compromised the Segnet to differentiate similar classes and deal with the intra-class variance. However, the end-to-end framework allows the image to be changed in order to help the Segnet, which makes it easier to differentiate similar classes. Also, considering that the LR aspect diminishes the intra-class variance, the results of the framework ended up being better than expected. 
 

\subsection{Comparison to LR training}

\begin{table}[h!]
\renewcommand{\arraystretch}{1.3}
\centering
\resizebox{\columnwidth}{!}{
\begin{tabular}{lllllll}
\hline
\textbf{Dataset}                    & \textbf{Deg.}         & \textbf{Method}  & \textbf{PSNR (dB)} & \textbf{Norm. acc}  & \textbf{IoU} & \textbf{Kappa}   \\
\hline
\multirow{4}{*}{Coffee}    & \multirow{2}{*}{$4\times$} & LR & 24.2429  & 0.7051 & 0.5591 & 0.4241   \\
                           &                     & End-to-end    &  26.055 & 0.8093 & 0.6807 & 0.6162  \\
                           \cline{2-7}
                           & \multirow{2}{*}{$8\times$} & LR &   25.7454  & 0.6356 & 0.4859 &  0.2984  \\
                           &                     & End-to-end    & 21.2722 & 0.7784  & 0.6477 & 0.5653\\
\hline
\multirow{4}{*}{Vaihingen} & \multirow{2}{*}{$4\times$} & LR & 28.7458 & 0.6020 & 0.5028  &  0.7116\\
                           &                     & End-to-end    & 26.3690   & 0.6631  & 0.5697  & 0.7738\\
                           \cline{2-7}
                           & \multirow{2}{*}{$8\times$} & LR & 25.3886 & 0.5756 & 0.4732  &  0.6849\\
                           &                     & End-to-end   & 22.6199   & 0.6625  & 0.5654  & 0.7730\\
\hline
\multirow{4}{*}{Thetford}      & \multirow{2}{*}{$4\times$} & LR & 26.8292 & 0.8146 & 0.6709 & 0.7824  \\
                           &                     & End-to-end    & 29.8188  & 0.8417  & 0.7117  & 0.7997\\
                           \cline{2-7}
                           & \multirow{2}{*}{$8\times$} & LR & 23.3354 & 0.8295 &  0.6385 & 0.7297 \\
                             &                     & End-to-end    & 25.5920 & 0.8564 & 0.6986 & 0.7881 
\\ \hline
\end{tabular}
}
\caption{Comparison between the performance of a Segnet trained only with LR data and the end-to-end framework.}
\label{tb:reconstruction}
\end{table}

Table~\ref{tb:reconstruction} presents the results for semantic segmentation by training with LR inputs and by using the end-to-end framework. The use of the framework improved the semantic segmentation results for all datasets, especially for the higher degradation factor ($8\times$).

Given that the framework presented higher semantic segmentation results than LR inputs, we can conclude that it is more accurate when predicting the labels for the pixels. For example, in the coffee dataset, the normalized accuracy was improved from $63\%$ in LR $8\times$ degradation to $77\%$ with the end-to-end framework. For the Vaihingen dataset, the same metric was increased from $57\%$ to $66\%$. This shows that the framework can indeed reconstruct LR inputs with visual details that make their pixel classes more discernible and accurately segmented.

Concerning the reconstruction, it is possible to see that the PSNR with the end-to-end framework is sometimes lower than the ones from interpolated LR images. However, in these cases, the Segnet results are better. That happens because the supervision of the semantic segmentation network in the training of the super-resolution method allows the framework to change the visual characteristics of the reconstructed image. This makes the PSNR drop, since the super-resolution output will present texture details that are nonexistent in the ground-truth HR image. But those details are exactly what makes the performance of the Segnet improve. An example of this situation can be seen in Figure~\ref{fig:exp2_visual}. Thus, we can conclude that lower PSNR values do not necessarily imply bad semantic segmentation performance. 

\subsection{Robustness to small object segmentation}

\begin{figure*}[h!]
  \begin{subfigure}
    \centering
    \includegraphics[width=0.35\textwidth]{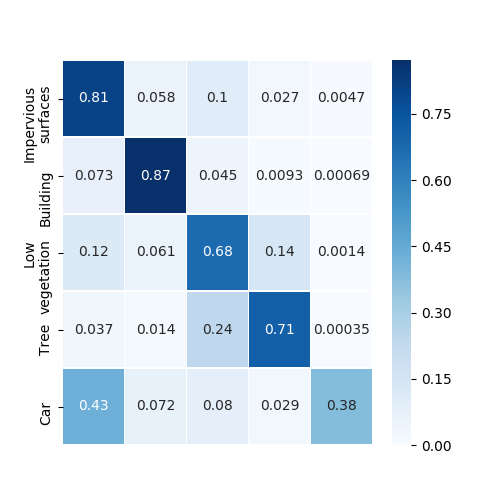}
  \end{subfigure}
 \begin{subfigure}
    \centering
    \includegraphics[width=0.35\textwidth]{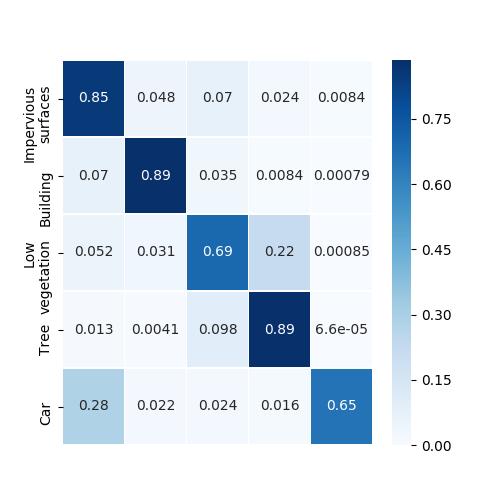}
  \end{subfigure}
  \begin{subfigure}
    \centering
    \includegraphics[width=0.35\textwidth]{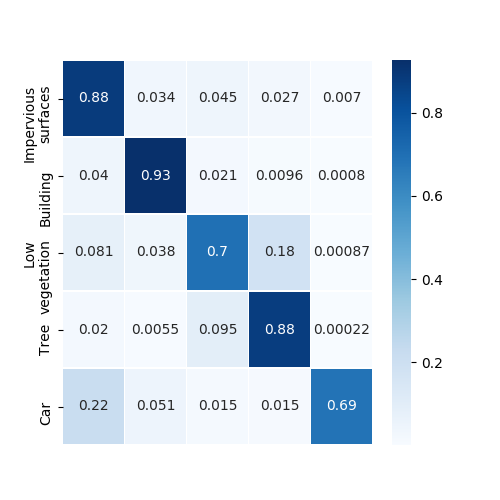}
  \end{subfigure}
 \caption{Confusion matrices for the Vaihingen dataset with $8\times$ degradation. From left to right: (i) Segnet trained with LR inputs, (ii) end-to-end framework, (iii) Segnet trained with HR inputs.}
 \label{fig:heatmap_vaihingen}
\end{figure*}

\begin{figure}[h!]
  \begin{subfigure}
    \centering
    \includegraphics[width=0.48\columnwidth]{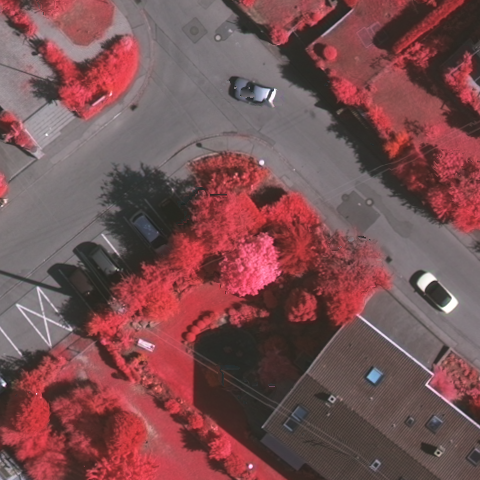}
  \end{subfigure}
 \begin{subfigure}
    \centering
    \includegraphics[width=0.48\columnwidth]{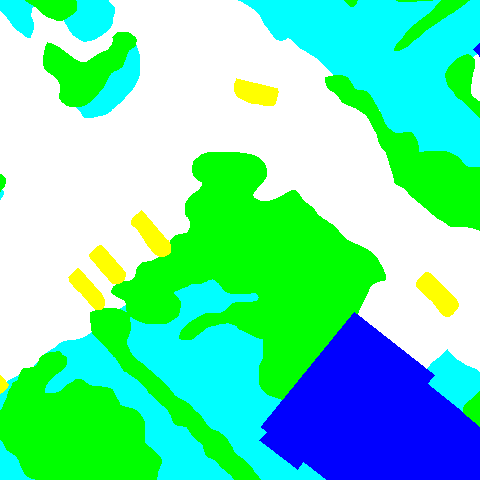}
  \end{subfigure}
  
  \begin{subfigure}
    \centering
    \includegraphics[width=0.48\columnwidth]{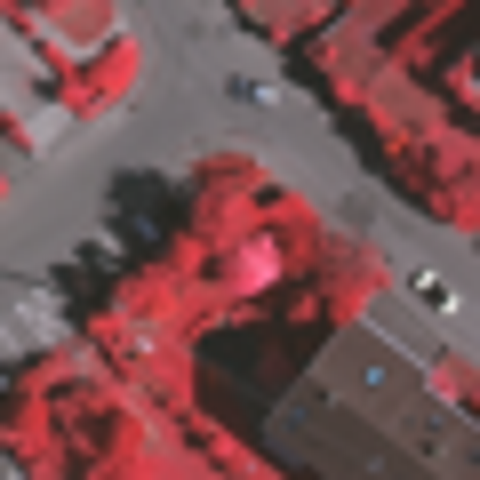}
  \end{subfigure}
 \begin{subfigure}
    \centering
    \includegraphics[width=0.48\columnwidth]{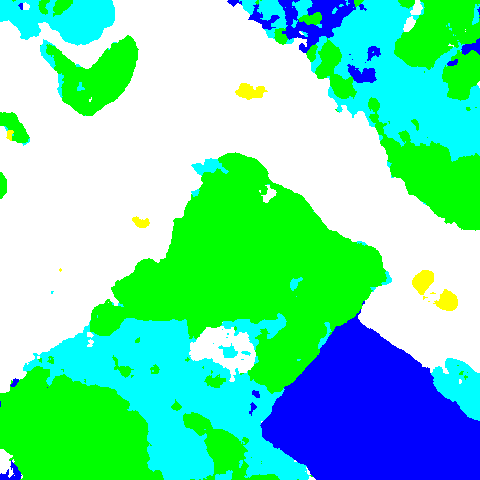}
  \end{subfigure}
  
  \begin{subfigure}
    \centering
    \includegraphics[width=0.48\columnwidth]{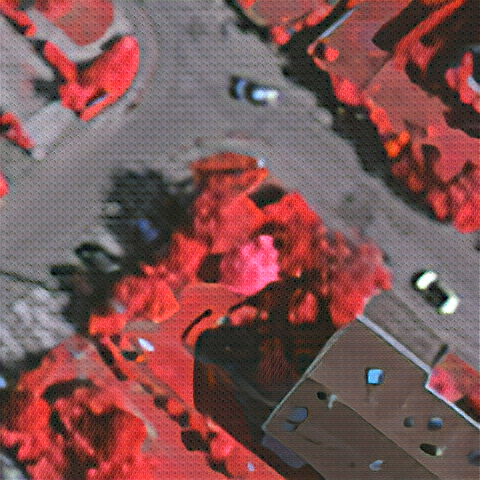}
  \end{subfigure}
 \begin{subfigure}
    \centering
    \includegraphics[width=0.48\columnwidth]{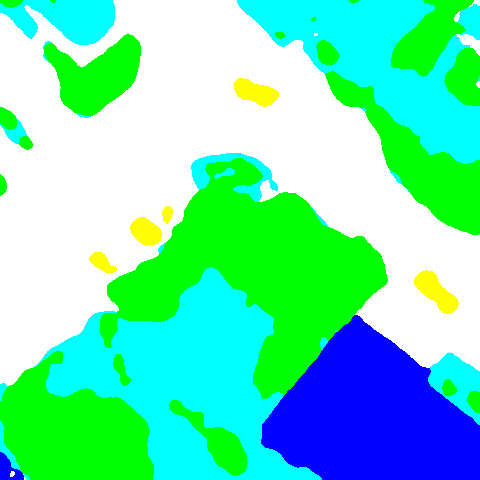}
  \end{subfigure}
 \caption{Visual example for the Vaihingen dataset. From up to bottom: (i) HR image and ground truth thematic map, (ii) LR image and its corresponding predicted map, (iii) end-to-end framework reconstructed image and its corresponding predicted map.}
 \label{fig:exp2_visual}
\end{figure}

Figure \ref{fig:heatmap_vaihingen} presents the confusion matrix for the Vaihingen dataset under three situations: Segnet trained with LR inputs, using the end-to-end framework, and Segnet trained with HR inputs. Visual results for super-resolution and semantic segmentation are shown in Figure \ref{fig:exp2_visual}.

Comparing the confusion matrices, we can see that the end-to-end framework managed to stay close to the accuracy of HR inputs. LR inputs, on the other hand, prejudiced the most the car class. In Figure~\ref{fig:exp2_visual} we can see an example of a segmentation that missed a lot of the car information due to the LR representation, but that was successfully recovered with the use of the framework. Looking at the confusion matrices, a Segnet trained with LR inputs predicted correctly only $38\%$ of the car pixels, while the framework achieved $65\%$ accuracy. The Segnet trained with HR data achieved $69\%$, which is not far from the framework's result. This confirms that the framework is capable of making more easily discernible objects that are small and difficult to see in an LR representation. The results also present a great improvement for the Tree class, which was better segmented on the framework. It increased the value from $71\%$ to $89\%$, even slightly better than HR inputs.

Finally, by observing the visual results of the reconstructed image from the end-to-end framework in Figure \ref{fig:exp2_visual}, it is possible to see (better viewed in a computer screen) the different textures created by the semantic segmentation network that helped it to classify the pixels more accurately. 

\section{Conclusion} \label{sec:conclusion}

In this paper, we proposed an end-to-end framework for semantic segmentation on LR remote sensing data. The framework trains a single network that shares the loss of both super-resolution and semantic segmentation. 

The framework improved the semantic segmentation performance for LR inputs
. The recovered textures created with the guide of the Segnet module greatly help it not to mislabel similar classes. These textures are artifacts that do not exist in the HR ground truth, which may lead to lower PSNR results, but without compromising the semantic segmentation performance. Furthermore, small objects, such as cars, can become more discernible with the framework, while not being easily detected in an LR space. The end-to-end framework also made it possible for LR inputs to achieve semantic segmentation results close to native HR data.


For future work, we plan to apply the proposed framework in real-world data (using images from more than one satellite with different resolutions, instead of manually down-sampled images), while using the other available bands during the training in order to improve the results even more. 

 \section*{ACKNOWLEDGEMENTS} \label{ACKNOWLEDGEMENTS}
The authors would like to thank NVIDIA for the donation of GPUs
. We also thank CAPES, CNPq, and FAPEMIG for the financial support provided for this research project.

{
    \begin{spacing}{1.17}
        \normalsize
        \bibliography{refs} 

\begin{thebibliography}{xx}

\bibitem[{Badrinarayanan} et al., 2017]{segnet}
{Badrinarayanan}, V., {Kendall}, A., {Cipolla}, R., 2017.
 SegNet: A Deep Convolutional Encoder-Decoder Architecture for Image
  Segmentation.
 {\em IEEE Transactions on Pattern Analysis and Machine Intelligence}, 39(12).

\bibitem[{Dai} et al., 2016]{dai2016}
{Dai}, D., {Wang}, Y., {Chen}, Y., {Van Gool}, L., 2016.
 Is image super-resolution helpful for other vision tasks?
 \emph{2016 IEEE Winter Conference on Applications of Computer Vision (WACV)},
  1--9.

\bibitem[Ferdous et al., 2019]{ferdous2019}
Ferdous, S.~N., Mostofa, M., Nasrabadi, N.~M., 2019.
 {Super resolution-assisted deep aerial vehicle detection }.
 T.~Pham (ed.), \emph{Artificial Intelligence and Machine Learning for
  Multi-Domain Operations Applications},  11006, International Society for
  Optics and Photonics, SPIE, 432 -- 443.

\bibitem[{Guo} et al., 2019]{guo2019}
{Guo}, Z., {Wu}, G., {Song}, X., {Yuan}, W., {Chen}, Q., {Zhang}, H., {Shi},
  X., {Xu}, M., {Xu}, Y., {Shibasaki}, R., {Shao}, X., 2019.
 Super-Resolution Integrated Building Semantic Segmentation for Multi-Source
  Remote Sensing Imagery.
 {\em IEEE Access}, 7, 99381-99397.

\bibitem[Haris et al., 2018a]{ddbpn}
Haris, M., Shakhnarovich, G., Ukita, N., 2018a.
 Deep backprojection networks for super-resolution.
 \emph{Conference on Computer Vision and Pattern Recognition (CVPR)}.

\bibitem[Haris et al., 2018b]{haris2018task}
Haris, M., Shakhnarovich, G., Ukita, N., 2018b.
 Task-driven super resolution: Object detection in low-resolution images.
 {\em arXiv preprint arXiv:1803.11316}.

\bibitem[Kim et al., 2016]{kim2016vdsr}
Kim, J., Kwon~Lee, J., Mu~Lee, K., 2016.
 Accurate image super-resolution using very deep convolutional networks.
 \emph{Proceedings of the IEEE Conference on Computer Vision and Pattern
  Recognition}, 1646--1654.

\bibitem[Ledig et al., 2017]{ledig2017}
Ledig, C., Theis, L., Husz{\'a}r, F., Caballero, J., Cunningham, A., Acosta,
  A., Aitken, A.~P., Tejani, A., Totz, J., Wang, Z. et~al., 2017.
 Photo-realistic single image super-resolution using a generative adversarial
  network.
 \emph{CVPR}, ~2number~3, 4.

\bibitem[Liu et al., 2016]{liu2016}
Liu, W., Anguelov, D., Erhan, D., Szegedy, C., Reed, S., Fu, C.-Y., Berg,
  A.~C., 2016.
 Ssd: Single shot multibox detector.
 \emph{Computer Vision -- ECCV 2016}, Springer International Publishing,
  21--37.

\bibitem[{Nogueira} et al., 2019]{nogueira2019}
{Nogueira}, K., {Dalla Mura}, M., {Chanussot}, J., {Schwartz}, W.~R., {dos
  Santos}, J.~A., 2019.
 Dynamic Multicontext Segmentation of Remote Sensing Images Based on
  Convolutional Networks.
 {\em IEEE Transactions on Geoscience and Remote Sensing}, 1-18.

\bibitem[Penatti et al., 2015]{coffee}
Penatti, O.~A., Nogueira, K., Dos~Santos, J.~A., 2015.
 Do deep features generalize from everyday objects to remote sensing and aerial
  scenes domains?
 \emph{Proceedings of the IEEE conference on computer vision and pattern
  recognition workshops}, 44--51.

\bibitem[Pereira, dos Santos, 2019]{sibgrapi}
Pereira, M.~B., dos Santos, J.~A., 2019.
 How effective is super-resolution to improve dense labelling of coarse
  resolution imagery?
 \emph{2019 32nd Conference on Graphics, Patterns and Images (SIBGRAPI)}, IEEE,
  202--209.

\bibitem[Pouliot et al., 2018]{pouliot2018}
Pouliot, D., Latifovic, R., Pasher, J., Duffe, J., 2018.
 Landsat Super-Resolution Enhancement Using Convolution Neural Networks and
  Sentinel-2 for Training.
 {\em Remote Sensing}, 10(3).

\bibitem[Ronneberger et al., 2015]{unet}
Ronneberger, O., Fischer, P., Brox, T., 2015.
 U-net: Convolutional networks for biomedical image segmentation.
 \emph{Medical Image Computing and Computer-Assisted Intervention -- MICCAI
  2015}, Springer International Publishing, 234--241.

\bibitem[Schowengerdt, 2006]{schowengerdt2006}
Schowengerdt, R.~A., 2006.
 {\em Remote sensing: models and methods for image processing}.
 Elsevier.

\bibitem[Shermeyer, Van~Etten, 2018]{shermeyer2018}
Shermeyer, J., Van~Etten, A., 2018.
 The Effects of Super-Resolution on Object Detection Performance in Satellite
  Imagery.
 {\em arXiv preprint arXiv:1812.04098}.

\bibitem[Shi et al., 2016]{espcn}
Shi, W., Caballero, J., Huszar, F., Totz, J., Aitken, A.~P., Bishop, R.,
  Rueckert, D., Wang, Z., 2016.
 Real-time single image and video super-resolution using an efficient sub-pixel
  convolutional neural network.
 \emph{The IEEE Conference on Computer Vision and Pattern Recognition (CVPR)}.

\end{thebibliography}
    \end{spacing}
}

\end{document}